\documentclass[10pt,twocolumn,letterpaper]{article}

\usepackage{3dv}
\usepackage{times}
\usepackage{epsfig}
\usepackage{graphicx}
\usepackage{amsmath}
\usepackage{amssymb}
\usepackage{booktabs}
\usepackage{enumitem}
\usepackage{multirow}
\usepackage{arydshln}
\usepackage{authblk}

\usepackage{color}
\usepackage{setspace}

\usepackage[pagebackref=true,breaklinks=true,letterpaper=true,colorlinks,bookmarks=false]{hyperref}

\threedvfinalcopy 

\def\threedvPaperID{70} 

\ifthreedvfinal\fi

\graphicspath{{figures/}}

\newcommand{\boldparagraph}[1]{\vspace{0.2em}\noindent{\bf #1} }

\let\oldnocite\nocite
\makeatletter
\renewcommand*{\nocite}[1]{\oldnocite{#1}\Hy@backout{#1}}
\makeatother

\DeclareMathOperator*{\minimize}{minimize}

\DeclareMathOperator*{\nDiv}{div}

\newcommand{\nR}{\mathbb{R}}                

\newcommand{\nSet}[1]{\{#1\}}
\newcommand{\nProjSet}{\Pi}			      

\newcommand{\nImg}{\mathrm{I}}              
\newcommand{\nTex}{\mathrm{T}}              

\newcommand{\nImgDom}{\Omega}               
\newcommand{\nTexDom}{\mathbb{T}}           

\newcommand{\nMesh}{\mathcal{M}}            

\newcommand{\nP}{\mathrm{P}}                
\newcommand{\nW}{\mathrm{W}}                

\newcommand{\nWeightG}{\mathrm{g}}          

\newcommand{\nK}{\mathrm{K}}                
\newcommand{\nKstd}{\mathbf{\sigma}}        
\newcommand{\nD}{\mathrm{D}}                

\newcommand{\nDualR}{\xi}                   
\newcommand{\nDualD}{\phi}                  
\newcommand{\nPDiter}{t}                    
\newcommand{\nNumPDiter}{t_\text{max}}      

\newcommand{\nNumCam}{N}                    

\newcommand{\nLoss}{\mathcal{L}}            

\begin{document}

\title{Learned Multi-View Texture Super-Resolution \vspace{-0.75cm}}

\author[1]{Audrey Richard}
\author[1]{Ian Cherabier}
\author[1]{Martin R. Oswald}

\author[2]{\authorcr Vagia Tsiminaki\thanks{{This work has been done while the author was at ETH Zurich.}}}
\author[1,3]{Marc Pollefeys}
\author[1]{Konrad Schindler}

\affil[1]{ETH Zurich, $^{2}$IBM Research Zurich, $^{3}$Microsoft Zurich}

\renewcommand\Authands{ and }
		
\maketitle

\ifthreedvfinal\thispagestyle{empty}\fi

\begin{abstract}
  We present a super-resolution method capable of creating a
  high-resolution texture map for a virtual 3D object from a set
  of lower-resolution images of that object. Our architecture
  unifies the concepts of (i) multi-view super-resolution based on the
  redundancy of overlapping views and (ii) single-view
  super-resolution based on a learned prior of high-resolution (HR) image
  structure. The principle of multi-view super-resolution is to invert
  the image formation process and recover the latent HR
  texture from multiple lower-resolution projections. We map that
  inverse problem into a block of suitably designed neural network
  layers, and combine it with a standard encoder-decoder network for
  learned single-image super-resolution. Wiring the image formation
  model into the network avoids having to learn perspective mapping
  from textures to images, and elegantly handles a varying number of
  input views. Experiments demonstrate that the combination
  of multi-view observations and learned prior yields improved
  texture maps.

\end{abstract}

\section{Introduction}

Capturing virtual 3D object models is one of the fundamental tasks of
computer vision. Movies, computer games, and all sorts of future
virtual and augmented reality applications require methods to create
visually realistic 3D content. Besides reconstructing the best
possible 3D geometry, an equally important, but perhaps less
appreciated step of that modeling process is to generate high-fidelity
surface texture.
However, the vast majority of image-based 3D reconstruction methods
ignores the texture component and merely stitches or blends pieces of
the input images to a texture map in a post-processing step, at the
resolution of the inputs, \eg,
\cite{Debevec-et-al-SIGGRAPH-1996,Bernardini-et-al-TVCG-2001,Eisemann-et-al-CGF-2008,Waechter-et-al-ECCV-14}.

\begin{figure}[t!]
  \scriptsize
  \centering
  \setlength{\tabcolsep}{0.5mm}
  \newcommand{\sz}{1.72cm}
  \begin{tabular}{ccccc}
    \rotatebox{90}{\hspace{0.2cm} Input image} &
   
    \includegraphics[height=\sz]{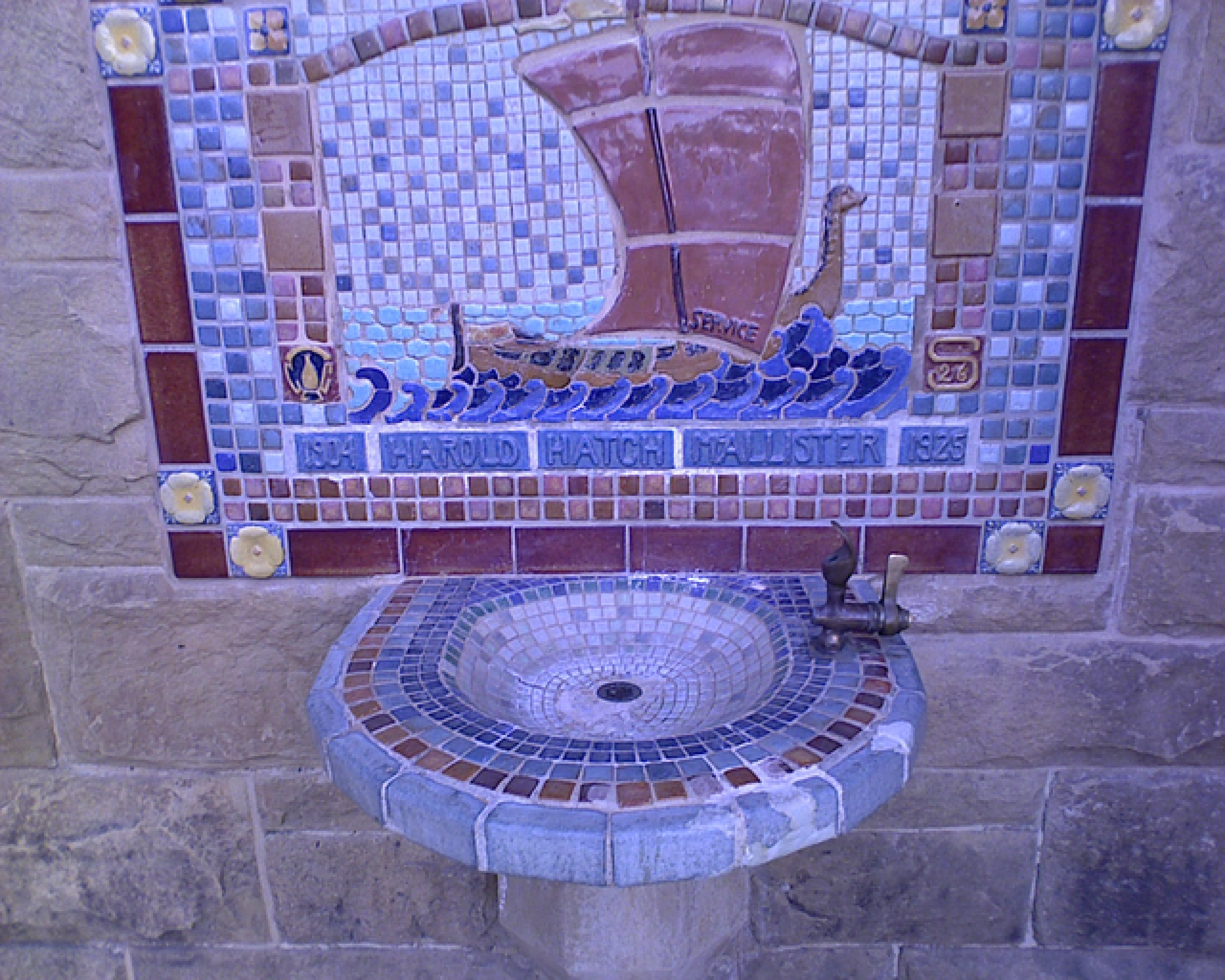} &    
    \includegraphics[height=\sz, trim={1136px 984px 1168px 824px},clip]{teaser_input_lr_view_11_x4.png} &
    \includegraphics[height=\sz, trim={1942px 1080px 538px 900px},clip]{teaser_input_lr_view_11_x4.png} &
    \includegraphics[height=\sz, trim={1934px 1224px 510px 700px},clip]{teaser_input_lr_view_11_x4.png} 
    \\
    %
    %
    \rotatebox{90}{\hspace{0.6cm} \cite{Tsiminaki-et-al-CVPR-2014}} &
    \includegraphics[height=\sz]{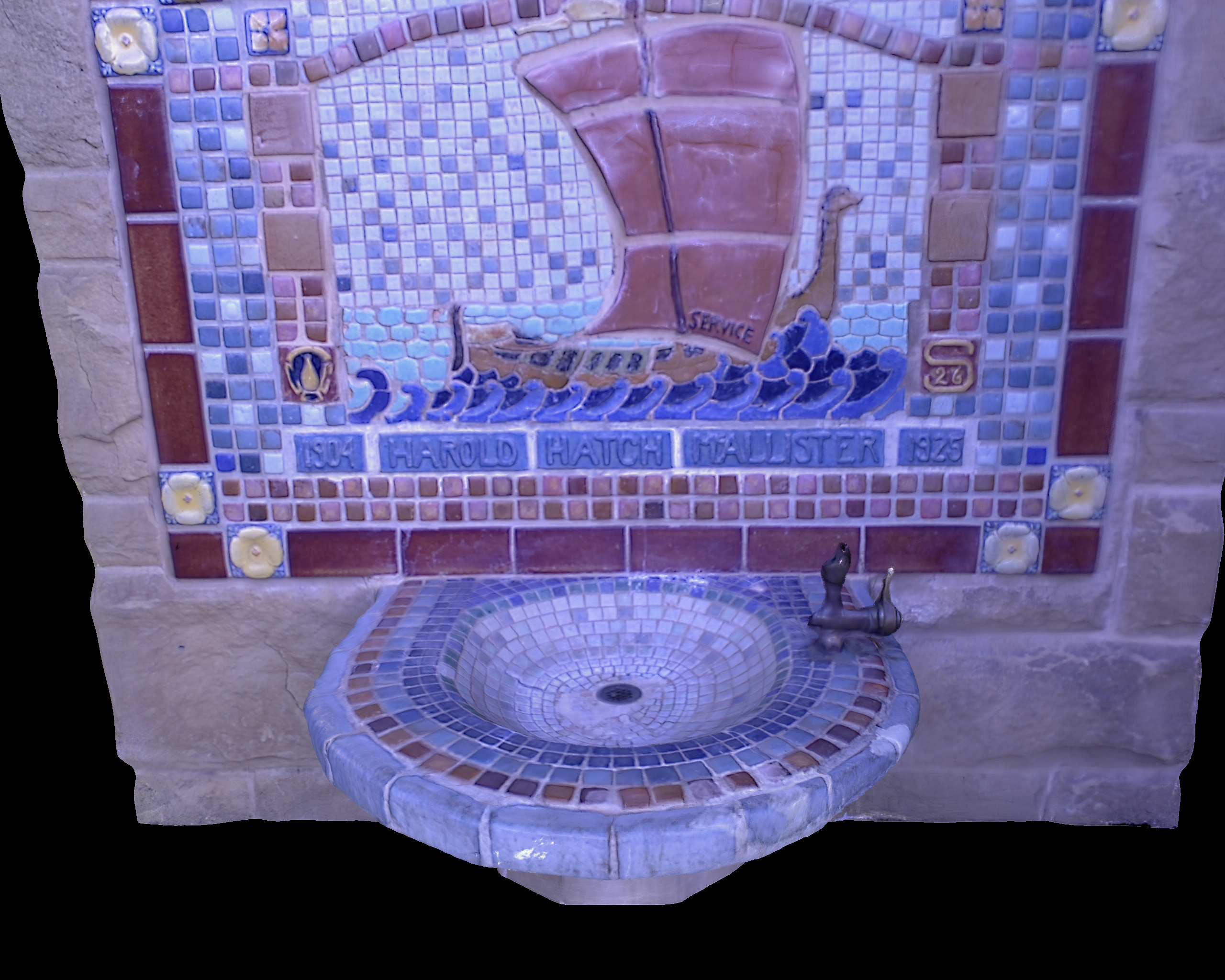} &
    \includegraphics[height=\sz, trim={1136px 984px 1168px 824px},clip]{teaser_tsiminaki.png}  &
    \includegraphics[height=\sz, trim={1939px 1078px 541px 902px},clip]{teaser_tsiminaki.png}  &   
    \includegraphics[height=\sz, trim={1923px 1220px 517px 704px},clip]{teaser_tsiminaki.png}  
    \\
    \rotatebox{90}{\hspace{0.6cm}Ours} &
    \includegraphics[height=\sz]{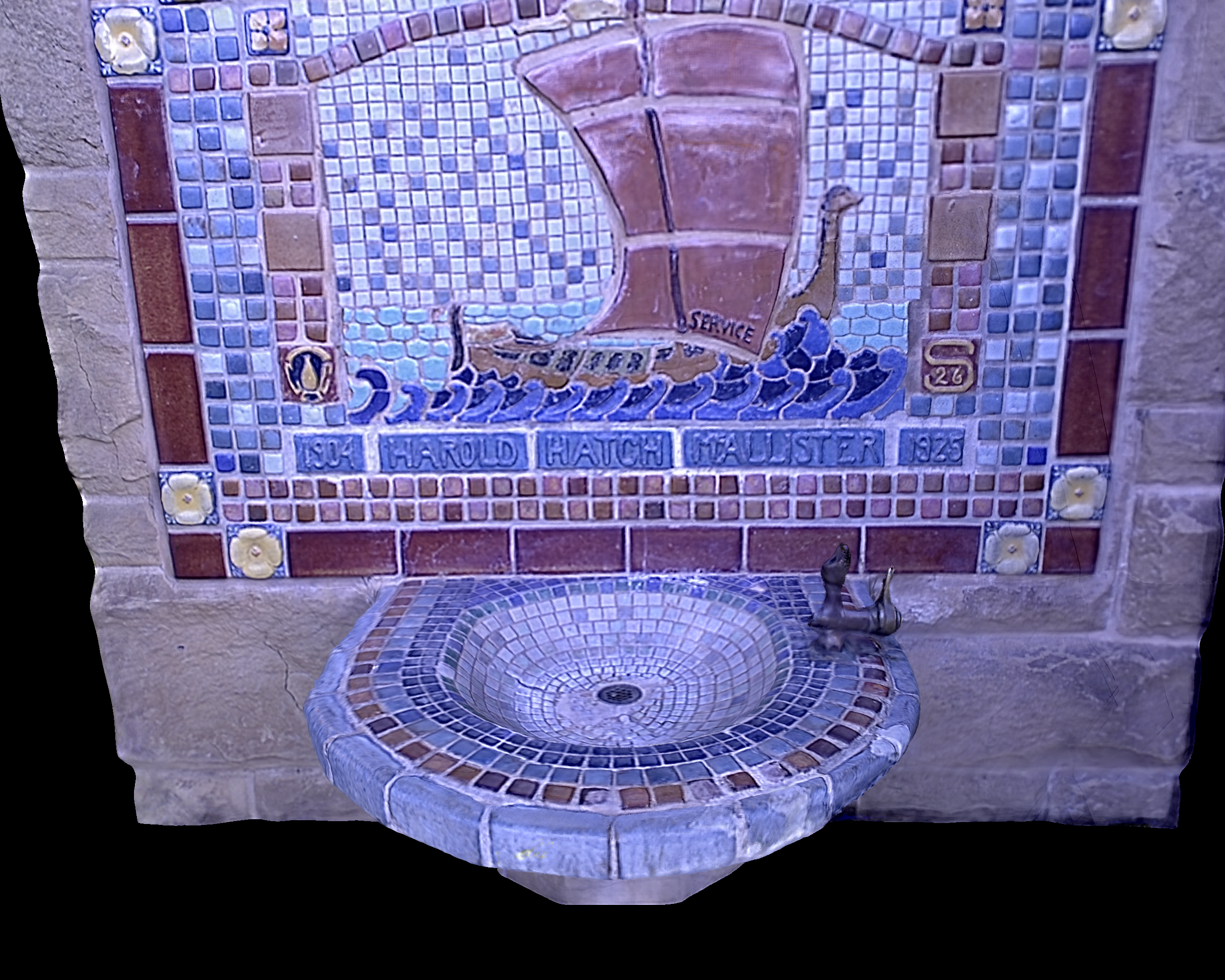} &
    \includegraphics[height=\sz, trim={1136px 984px 1168px 824px},clip]{teaser_ours_sequential_arch_10views.png} &
    \includegraphics[height=\sz, trim={1939px 1078px 541px 902px},clip]{teaser_ours_sequential_arch_10views.png} &
    \includegraphics[height=\sz, trim={1923px 1220px 517px 704px},clip]{teaser_ours_sequential_arch_10views.png}
  \end{tabular}
  \caption{Learned super-resolution result compared to the state-of-the-art (upscaling $\times 4$). The result of our learning-based multi-view method is significantly sharper than prior art~\cite{Tsiminaki-et-al-CVPR-2014}.}
  
  \label{fig:teaser}
  \vspace{-0.5cm}
\end{figure}

Here, we explore the possibility to compute a \emph{higher resolution
  texture map}, given a set of images with known camera
poses that observe the same 3D object, and a 3D surface mesh of the
object (which may or may not have been created from those images).  We
argue that sticking to the original resolution actually under-exploits
the image information: it is well established in the literature that,
under reasonable conditions, super-resolution (SR) by a factor of at least
$\times2$ to $\times4$ is feasible.  Figure~\ref{fig:teaser} shows an
example of our method that illustrates this task.


There are two fundamentally different approaches to image
super-resolution%
%
: \emph{(i)~redundancy-based multi-image SR},
which uses the fact that each camera view represents a different
spatial sampling of the same object surface. The resulting
oversampling can be used to reconstruct the underlying surface
reflectance at higher resolution in a physically consistent manner, by
inverting the image formation process that mapped the surface to the different
views.
On the other hand,
\emph{(ii)~prior-based single-image SR} 
aims to generate a plausible, visually convincing high-resolution (HR)
image from a single low-resolution (LR) image, by learning from
examples what HR patterns are likely to have produced the
low-resolution image content.  Also for this obviously ill-posed
problem, impressive visual quality has been achieved, particularly
with recent deep learning approaches. Clearly, single-image SR can
only ever ``hallucinate from memory'', since it is entirely based on
prior knowledge. There is no redundancy to constrain the reconstructed
high-frequency content to be physically correct.  In our
classification prior-based methods include recent learning-based
single-image SR techniques, but also techniques not based on learning,
\eg, exemplar-based ones.  Historically, the prior on the output
image has often been handcrafted, while today it is mostly learned
from training data.

Each of the two approaches has been shown -- separately -- to work
rather well on natural images.
But the vast majority of existing work is limited to either one or the
other.
Our goal in this paper is to unify them both into an integrated
computational model that combines their respective
advantages. 
Moreover, our model can learn to rely more on one or the other
paradigm, depending on the available data.
To our knowledge our work is the first such integrated model for the
general multi-view setting. 

In summary, the contributions of the present paper are:
\emph{(i)}~the first super-resolution framework capable of combining,
in a general multi-view setting, redundancy-based multi-view SR with
single-image SR based on a learned HR image prior.
\emph{(ii)}~A network architecture that merges state-of-the-art deep
  learning and traditional variational SR methods.
This unifying architecture has multiple advantages:
  \emph{(i)}~it seamlessly handles an arbitrary number of input
  images, and is invariant to their ordering; including the special
  case of a single image (falling back to pure single-view
  SR). 

  \emph{(ii)}~It does not waste resources, potentially sacrificing
  robustness, to learn known operations such as perspective projection
  to relate images taken from different viewpoints.
  \emph{(iii)}~It focusses the learning effort on small residual
  corrections, in both the single- and multi-view branches, thus reducing
  the amount of training data needed.

\section{Related Work}

In the following we focus on the most related works and group them
according to the problem setting. We differentiate between
the two very different paradigms of \textit{prior-based single-image SR} and
\textit{redundancy-based multi-image SR}.

\boldparagraph{Prior-based Single-image SR.}
%
The goal of single-image SR is to fill in HR image patterns by
leveraging a prior derived either from similar patches in other parts
of the input image (self-examplars)
\cite{Glasner-et-al-ICCV-2009,Huang-et-al-CVPR-2015}, or from similar
image patches from an existing image database
\cite{Freeman-et-al-CGA-2002}, or -- most commonly -- from previously
seen training data \cite{Kim2010SingleImageSU,
  NaiveBayesSuper-ResolutionForest,Timofte2013AnchoredNR,timofte2014a+}.
Recently the technologies of choice for learning the prior have been
(deep) convolutional networks
\cite{Dong-et-al-TPAMI-2016,Yamanaka-et-al-ICONIP-2017,Sajjad-et-al-ICCV-2017,Hui-et-al-CVPR-2018,Zhang-et-al-CVPR-2018,Shocher-et-al-CVPR-2018}
and generative adversarial
networks~\cite{Ledig-et-al-CVPR-2017,Wang-et-al-ECCVW-2018,Park-et-al-ECCV-2018,Wang-et-al-CVPRW-2018,Wang-et-al-CVPR-2019}.
%
%
%
An overview of recent single-image SR methods is given in the report
of the NTIRE Challenge~\cite{Ancuti-et-al-CVPRW-2018}. Further surveys
can be found in
\cite{Hayat-ArXiV-2017,Yang-et-al-ArXiv-2018,Wang-et-al-ArXiv-2019}.
%
%
%
%
Single-view SR has also been applied to the 3D setting, for the special
case of face reconstruction, to generate both texture maps
\cite{Saito-et-al-CVPR-2017} and displacement maps
\cite{Huynh-et-al-CVPR-2018}.


Single-image SR methods upsample to some ``educated
guess''. They inpaint/hallucinate high-frequency information
that is \emph{plausible} according to previously seen examples, but
they intrinsically cannot distinguish between the true HR
image and another plausible candidate.

\boldparagraph{Redundancy-based Multi-image SR.}
%
In contrast to single-image SR, multi-image SR recovers
high-frequency details from the redundant observations afforded by
multiple images that depict the same scene. The crucial condition
(which luckily is easier to meet than to violate in practice) is that
each image is captured with a small, sub-pixel offset relative to all
others, so that the surface is oversampled.

Multi-frame (including video) SR methods have been tackled with a
Bayesian model \cite{Liu-Sun-TPAMI-2014}, with variational methods
\cite{Mitzel-et-al-DAGM-2009,Unger-et-al-DAGM-2010}, and with deep
learning
\cite{Liao-et-al-ICCV-2015,Tao-et-al-ICCV-2017,Sajjadi-et-al-CVPR-2018,Jo-et-al-CVPR-2018}.
Also hybrid methods have been developed that build on the variational
approach, but learn the regularization \cite{Klatzer-et-al-GCPR-2017}.
In contrast to the multi-view setting considered in the present paper,
all these works require very small view point changes as well as a fixed
number of input frames.

\boldparagraph{Multi-view Texture Mapping.}
In order to compute the color of a surface point visible in multiple images,
one can either try to select the best view or blend multiple views.
Blending can in the simplest case mean averaging, typically weighted
according to visibility and surface viewing
angle~\cite{Debevec-et-al-SIGGRAPH-1996}.
To avoid over-smoothed textures, blur and ghosting artifacts, several
authors have introduced additional texture registration steps
\cite{Lensch-et-al-GM-2001,Bernardini-et-al-TVCG-2001,Theobalt-et-al-TVCG-2007,Eisemann-et-al-CGF-2008,Waechter-et-al-ECCV-14}.
Rather than blending multiple images, only one image was used in \cite{Lempitsky-Ivanov-CVPR-2007} as texture at any given location followed by a MRF model to avoid artifacts at seams.
A recurrent issue for texture generation is misregistration due to
inaccurate 3D geometry and/or camera calibration.
One can try to improve the geometry during texture
mapping~\cite{Takai-et-al-3DPVT-2010} for better texture quality.
A more common strategy is to estimate a generic 2D optical flow field
per image that compensates small registration errors
\cite{Eisemann-et-al-CGF-2008,Waechter-et-al-ECCV-14}.
%
%
Overall, the goal of these methods is to prevent degradations during
texturing, their upper bound is the quality of an individual source
image. Texture SR aims to exceed the quality of individual input
images.

\boldparagraph{Multi-view Texture SR.}
Early work~\cite{Koch-et-al-ECCV-1998} suggested to create textures
with a resolution higher than the input images through oversampling.
The seminal work \cite{Goldluecke-Cremers-DAGM-2009} was perhaps the
first that really super-resolved texture maps.
That line of work was later extended to also refine surface geometry
\cite{Goldluecke-Cremers-ICCV-2009} and camera calibration
\cite{Goldluecke-et-al-IJCV-2014}.
Also here, it was shown that calibration and geometry errors can be
compensated more effectively with the help of optical
flow~\cite{Tsiminaki-et-al-CVPR-2014}.
%
In \cite{Delaunoy-Pollefeys-CVPR-2014}, the authors focus on surface
refinement, but also perform texture super-resolution, with a simple
bilinear kernel.

An interesting, albeit computationally demanding alternative is to
implicitly super-resolve the texture by reconstructing surface
elements with constant colour, but very high-resolution, as
in~\cite{Maier-et-al-ICCV-2017}. There abundant RGB-D data is
integrated into a HR voxel model. With very accurate
geometry and calibration, texture can then be computed by simple color
blending per surface voxel.

All these works exploit view redundancy, while ignoring the
possibility to learn an a-priori model of HR image
statistics, as learning-based single-view methods do.


\boldparagraph{Multi-view Learning-based Texture SR.}
Only recently, deep learning-based methods have been introduced in the
multi-view case to enhance the quality of the reconstructed models,
\eg, for facial scans~\cite{nhan2015beyond}.
%
For facial animation a deep generative network to infer per-frame texture deformation was introduced~\cite{olszewski2017realistic}.
In the context of cloth simulation one can compute a global shape deformation and generate high-frequency normal maps with a GAN~\cite{lahner2018deepwrinkles}. 
Recently, per-image SR was extended to the multi-view case, by
injecting the 3D information in the form of normal
maps~\cite{Li_2019_CVPR}, reaching similar performance as 3D
model-based methods.

The latter works leverage the capacity of deep neural networks, yet a
method that explicitly incorporates both model-based and
learning-based techniques is still missing.  In this paper, we
construct such a method, by converting the model-based pipeline into a
neural network architecture and merging it with conventional, learned
single-image SR.

\section{Method}
\label{sec:method}

\boldparagraph{Overview.}
The inputs to our method are a set of calibrated LR images and a 3D surface given by a mesh.
Its output is a SR texture atlas that can be directly used for 3D rendering.

\begin{figure*}[ht!]
  \includegraphics[width=\linewidth]{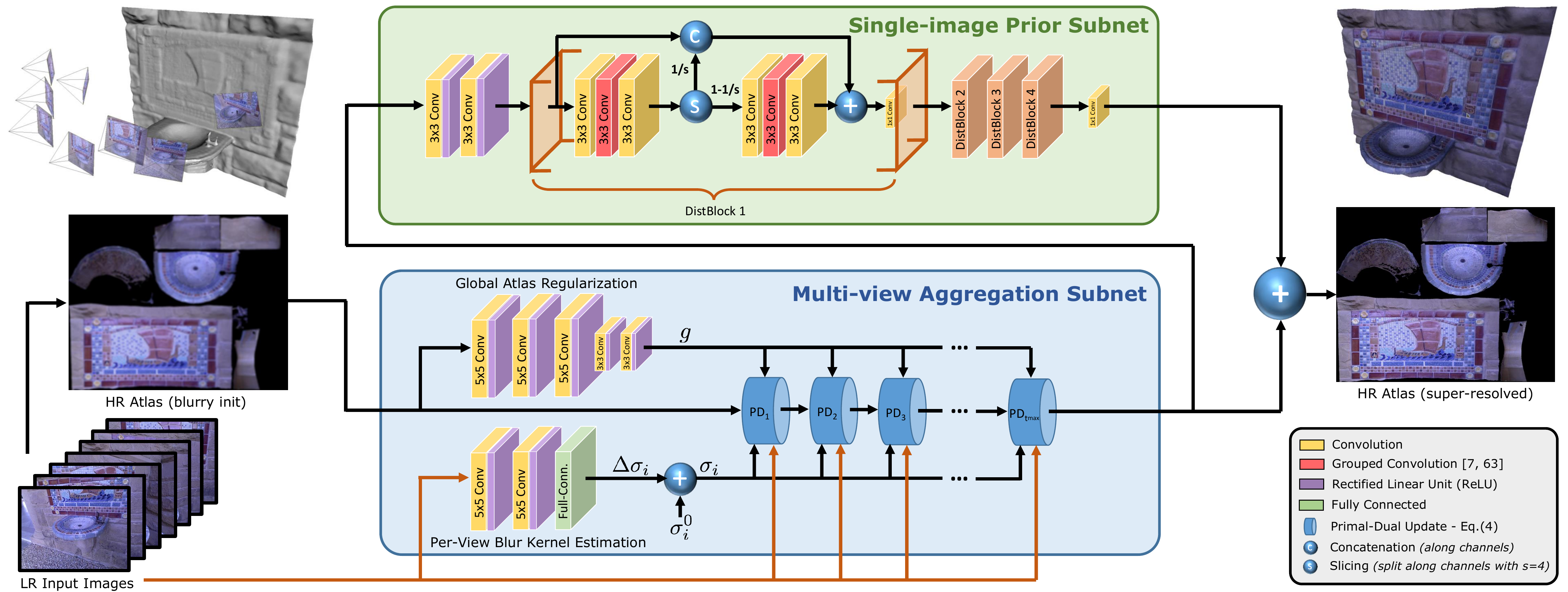}
  \caption{The proposed multi-view super-resolution network combines
    the concepts of \emph{redundancy-based multi-view SR (blue)} and
    \textit{prior-based single-image SR (green)} in a single
    end-to-end trainable architecture.  The \emph{multi-view
      aggregation (MVA) subnet} leverages the view redundancy to
    construct an intermediate SR texture atlas. The
    \emph{single-image prior (SIP) subnet} learns the differences
    between this atlas and the ground truth to further boost the SR
    performance.  }
  \label{fig:network_architecture}
\end{figure*}

Using the calibrated LR images, we estimate an initial (still blurry) texture atlas.
The way the texture atlas is created is arbitrary.
We then use a neural network architecture consisting of two major parts to super-resolve the texture atlas.
The \textit{redundancy-based} part aggregates multi-view redundancy from all LR input images and computes an intermediate, super-resolved texture atlas. 
The \textit{prior-based} part enhances the single, intermediate atlas using knowledge about its expected statistics that it has collected from a training dataset.
We refer to the former as Multi-View Aggregation (MVA) subnet, and the latter as Single-Image Prior (SIP) subnet.
The overall network architecture with the two sub-networks is shown in Figure~\ref{fig:network_architecture} and the individual parts are detailed in the following subsections.
%

\subsection{Multi-View Aggregation (MVA)}\label{sec:MVA}
%
The goal of MVA is to generate a super-resolved texture atlas by leveraging multi-view redundancy.
The corresponding network branch is shown in the blue box of Figure~\ref{fig:network_architecture}.
It exploits the image formation model to explicitly define the texture-to-input correspondence via projections.

\boldparagraph{Motivation.}
In a multi-view setting the number of views that are useful for reconstructing the texture at a particular surface point varies, as it depends on the camera setup and the surface geometry (including self-occlusions).
To deal with varying number of inputs, deep networks typically use pooling~\cite{Ranzato-et-al-CVPR-2007,Aittala-Durand-ECCV-2018} or recurrent architectures (\eg, LSTMs~\cite{Hochreiter-Schmidhuber-NC-1997}).
We argue that none of them is well-suited for redundancy-based SR:
the \emph{average pooling} operation blurs multiple inputs into a single value, which is counter-productive when aiming to recover HR information. 
\emph{Max-pooling} selects only a single input, discarding multi-view information -- information mixed by integration cannot be recovered by selection.
As for \emph{recurrent networks}, their result depends on the order in which the images are processed, which contradicts common sense (and the image formation model).
Instead, we propose to incorporate the (inverse) image formation model into the network in the form of special layers that perform redundancy-based multi-image SR. By explicitly modeling blurring and downsampling, one can relate any number of LR input views to the desired HR output.
%

\boldparagraph{Notation.}
We are given a set of $\nNumCam$ LR images $\nSet{\nImg_i}_{i=1}^{\nNumCam}$, with $\nImg_i\!: \nImgDom_i \subset \nR^2 \rightarrow \nR$ (grayscale).
The scene geometry is provided as a mesh $\nMesh$.
For each LR image, we precompute a projection matrix that maps the texture domain to the image domain through $\nMesh$ via $\nP_i\!: \nR^2 \rightarrow \nR^2$,
following \cite{Tsiminaki-et-al-CVPR-2014}. 
The output SR texture atlas is denoted $\nTex:\nTexDom \subset \nR^2 \rightarrow \nR$.

\boldparagraph{Image Formation Model.}
Each input image depicts the same 3D geometry from a different viewpoint, leading to a different sampling of the surface.
Together, they oversample the surface, which makes it possible to reconstruct its reflected color at higher resolution.
In order to recover that HR signal, we aim to invert the image formation process.
Namely, the camera lens blurs the incoming bundle of rays, and the sensor integrates (averages) the contribution of all incoming photons within a pixel. 
Following~\cite{Elad-Feuer-TPAMI-1999}, we have:
\begin{equation}
  \nImg_i = \nD\nK_i\nW_i\nP_i \cdot \nTex + e_i 
  \label{eq:image_formation}
\end{equation}
An LR input image $\nImg_i$ is generated from the SR texture $\nTex$ via a sequence of perspective projection $\nP_i$, blurring with convolution kernel $\nK_i$, downsampling with operator $\nD$ and adding noise $e_i$.
The additional warp operator $\nW_i$ accounts for geometric inaccuracies and camera calibration errors with an optical flow correction, precomputed as in \cite{Waechter-et-al-ECCV-14, Tsiminaki-et-al-CVPR-2014}.

\boldparagraph{SR Multi-view Energy.}
Similar to \cite{Goldluecke-Cremers-ICCV-2009,Tsiminaki-et-al-CVPR-2014} we compute an SR texture atlas $\nTex$ as a minimizer of the following energy: %
\begin{equation}
  \minimize_\nTex  \;\; \sum_{i=1}^{\nNumCam} 
    \left\| \nD\nK_i\nW_i\nP_i \cdot \nTex - \nImg_i \right\|_1   
    + \|\nWeightG \cdot \nabla \nTex\|_1
    \label{eq:energy}
\end{equation}
The energy minimizes the re-projection error between the SR texture atlas and all inputs $\nSet{\nImg_i}_{i=1}^\nNumCam$.
The second term is the weighted total variation regularizer, with $\nWeightG$ a locally-adaptive weight function.
It favors a piecewise constant solution if the input data is ambiguous or absent and is necessary to make the optimization well-posed. 

\boldparagraph{Parameter Estimation.}
In contrast to classical SR approaches \cite{Goldluecke-Cremers-DAGM-2009,Tsiminaki-et-al-CVPR-2014} for which $\nWeightG$ is manually set to some constant, we locally estimate the amount of smoothing from the texture atlas with a CNN, such that it yields the optimal SR reconstruction. 
Similarly, another small network adjusts the standard deviation $\sigma_i$ of the Gaussian blur kernel $\nK_i$ for each view $i$. See Figure~\ref{fig:network_architecture} (left side of blue box).

\boldparagraph{MVA Layers.}
For numerical optimization of ~\eqref{eq:energy} we employ a first-order primal-dual method \cite{Chambolle-Pock-JMIV-2011}, known to be a particularly suitable solver for this type of problems.
To that end, the dual variables $\nDualD, \nDualR$ are introduced via the Legendre-Fenchel transform of the regularizer, in order to deal with the non-differentiability of the $L^1$-norm. 
This transforms ~\eqref{eq:energy} into the following saddle-point problem:
\begin{equation}
  \min_\nTex \max_{\substack{\|\nDualD_i\|_\infty \leq 1 \\ \|\nDualR\|_\infty \leq 1 }} \;
  \sum_{i=1}^{\nNumCam}
    \langle \nD\nK_i\nW_i\nP_i \cdot \nTex - \nImg_i,  \nDualD_i \rangle
  %
  + \langle \nWeightG \cdot \nabla\nTex, \nDualR \rangle
  \label{eq:pd_energy}
\end{equation}
The algorithm in \cite{Chambolle-Pock-JMIV-2011} jointly performs a gradient descent in the primal variable $\nTex$ and a gradient ascent in the dual variables $\nDualD_i, \nDualR$ 
and iterates $\nPDiter\in\nSet{1,\ldots,\nNumPDiter}$ steps as follows:
\begin{subequations}
\label{eq:pd_updates}
\begin{align}
  \nDualD_i^{\nPDiter+1} &= \nProjSet_{\|\cdot\|\leq 1} 
  \left[ 
    \nDualD_i^\nPDiter + \eta \;(\nD\nK_i\nW_i\nP_i \bar{\nTex}^\nPDiter - \nImg_i)
    \right] \label{eq:pd_dual_update}\\
  \nDualR^{\nPDiter+1} &= \nProjSet_{\|\cdot\|\leq 1} 
    \left[ \nDualR^\nPDiter + \eta \cdot \nWeightG \cdot\nabla\bar{\nTex}^\nPDiter \right] \label{eq:pd_dual_update}\\ 
  \nTex^{\nPDiter+1} &= \nTex^\nPDiter + \tau \big(
    \nWeightG \cdot \nDiv \nDualR^{\nPDiter+1} - 
    \sum\nolimits_{i=1}^{\nNumCam} \nP_i^\intercal \nW_i^\intercal 
    \nK_i^\intercal \nD^\intercal \nDualD_i^{\nPDiter+1}
    \big) 
    \label{eq:pd_prim_update}\\
  \bar{\nTex}^{\nPDiter+1} &= 2\nTex^{\nPDiter+1} - \nTex^{\nPDiter}
  \label{eq:pd_overrelaxation}
\end{align}
\end{subequations}
where $\eta = \tau = 0.025$ are step sizes and $\nProjSet_{\|\cdot\|\leq 1}$ a projection onto the $L^2$ unit ball per pixel. 
Details can be found in the \textit{Supplementary Material}.
In order to transform the numerical minimization into a neural network, we unroll a fixed number of $\nNumPDiter$ optimization steps, such that each update cycle \eqref{eq:pd_updates} represents one network layer in our MVA subnet.
Unrolling is a generic technique to include iterative energy minimization into neural network blocks, used also in low-level vision \cite{Vogel-Pock-GCPR-2017}, medical image reconstruction \cite{Kobler-et-al-GCPR-2017}, single image depth super-resolution \cite{Riegler-et-al-ECCV-2016}, and semantic 3D reconstruction \cite{Cherabier-et-al-ECCV-2018}.
It allows us to combine the classical multi-view SR with learning-based single-view SR methods.
The updates~\eqref{eq:pd_updates} are matrix operations and can be regarded as specifically tailored convolutions that allow the multi-view aggregation.
The steps are depicted as blue cylinders in Figure~\ref{fig:network_architecture}, where each layer $PD_i$ represents a single primal-dual update~\eqref{eq:pd_updates}.
The SR texture atlas estimated from multiple views is then fed into the subsequent SIP subnet.
%

%
%
\nocite{Chollet_2017_CVPR}
\nocite{Xie_2017_CVPR}

\subsection{Single-Image Prior (SIP)}
%
The goal of the SIP subnet is to learn which high-frequency texture statistics typically underlie a given low-frequency pattern. 
Following recent single-image SR \cite{Hui-et-al-CVPR-2018}, we build our network in a way that the image prior is learned as a residual correction to the input image, since these high-frequency differences are easier to regress than the full image content.
However, instead of generic (bilinear or bicubic) upsampling, we can already provide the prediction of redundancy-based MVA as input.
For learning the residual prior, we follow the approach of the information distillation network \cite{Hui-et-al-CVPR-2018}, which exhibits a simple feed-forward network structure that also learns residual changes in composition with standard bicubic upsampling.
This state-of-the-art single-image SR architecture is simple and lightweight and can easily be adapted to be combined with our MVA network (see Section~\ref{sec:implementation}). 
The SIP subnet is depicted in the green box of Figure~\ref{fig:network_architecture}.

\boldparagraph{Note on Prior-based SR Correctness.}
The result of prior-based SR can only be \emph{plausible}, but effectively is hallucination based entirely on \emph{other} images.
This may be undesired for applications in which correctness is more important than visual quality. 
Such scenarios, by definition, rule out single-frame SR.
In contrast, our MVA network can also be used without the SIP part, and will likely outperform the classical energy minimization it is based on, due to its ability to learn the algorithm (meta-)parameters from data.

\subsection{Loss Function}
%
The overall network, as depicted in Figure~\ref{fig:network_architecture}, is trained in a supervised fashion.
Our main training objective is the $L^1$-norm of the intensity differences between the predicted texture atlas $\hat{\nTex}$ and the corresponding ground truth $\nTex$.
A second loss term is added that keeps 
the standard deviation $\nKstd_i$ of the blur kernel $\nK_i$ near the initial value $\nKstd_i^{0}$:
\begin{equation}
	\nLoss = \|\nTex - \hat{\nTex}\|_1 + 
		\alpha \cdot \sum_i \|\nKstd_i - \nKstd_i^{0}\|_2
\end{equation}
where $\alpha = 1$ scales the contribution of the two terms. 
Since the weighted TV-regularizer is defined on the texture gradient $\nabla T$,
predicting the weight function $\nWeightG$ is accounted for in the first term and a dedicated loss is not needed.

\section{Experiments}
\label{sec:experiments}

\subsection{Datasets}
%
\noindent
We use the following two datasets of different types:
\begin{itemize}[topsep=1pt,leftmargin=*]
  \setlength\itemsep{-1mm}   
   \item \textbf{Single-image DIV2K dataset \cite{Agustsson_2017_CVPR_Workshops} :} composed of 800 diverse 2K resolution high-quality natural images. The corresponding LR images $\times2$ and $\times4$ have been obtained synthetically using \emph{undisclosed} degradation operators.  
   \item \textbf{Multi-view 3D dataset:} following \cite{Goldluecke-Cremers-DAGM-2009,Goldluecke-et-al-IJCV-2014,Tsiminaki-et-al-CVPR-2014} we use 8 scenes (\ie, \textit{Beethoven}, \textit{Bird}, \textit{Bunny}, \textit{Fountain}, \textit{Buddha}, \textit{Head}, \textit{Relief}, \textit{Temple Ring}) including calibrated LR images, appearance projections, and 3D geometry.   
\end{itemize}
\noindent
We convert all data into YCbCr color space and only process the Y-channel, which contains pratically all high-frequency information.
We use bicubic upsampling of the CbCr-channels to reconstruct the super-resolved RGB output.
The DIV2K dataset is only used to pre-train the SIP subnet.
Contrary to single-image SR of natural images, our method requires a precomputed texture mask to delimit the different charts.
To increase the amount of data we perform data augmentation: \emph{(i)}~rotations with an angle of 90$^\circ$, 180$^\circ$, 270$^\circ$, \emph{(ii)}~horizontal and vertical flips, \emph{(iii)}~four random multiplicative brightness changes with random offsets. 
To deal with uninformative background in texture maps, we dilate the textured regions, such that the receptive field for a valid texture pixel contains no background.
Our main multi-view 3D dataset is then split into mutually exclusive testing (\textit{Beethoven}, \textit{Bird}, \textit{Bunny}) and training scenes (all others).
%

\subsection{Implementation Details}\label{sec:implementation}
%
\boldparagraph{Patch-based Approach.}
Working with real multi-view 3D data in combination with a neural network makes the implementation more challenging.
In order to handle the massive quantity of data, mostly due to the presence of the projections $P_i$ and the optical flows $\nW_i$, we adopt a \textit{patch-based approach} during training.
Input atlases and their corresponding super-resolved ground truths are cut into patches of size $64\!\times\!64$.
For every texture patch, we generate a corresponding patch in the
image domain, with a size of $200\!\times\!200$ to ensure the reprojection
falls completely inside the patch in spite of possible projective
distortions - as well as the corresponding optical flow.
Accordingly, the LR input images are divided into patches of size $100\!\times\!100$
(or $50\!\times\!50$) in order to train for an upsampling factor
$\times2$ (or $\times 4$).
%

\boldparagraph{MVA Subnet.}
During training, we limit the maximum number of LR views to $N = 20$. 
While conceptually an arbitrary number is possible, having the same number greatly simplifies batch processing. 
We thus only use 20 views, the minimum number available for any scene in the multi-view dataset.
The selected views fully cover each object, with the best possible overlap for multi-view SR.
To obtain an initial atlas (not yet super-resolved), we follow \cite{Tsiminaki-et-al-CVPR-2014} and compute visibility masks from texture to image domain. 
They are used to generate the projection operators, as well as the initial atlas averaged from corresponding visible input views.
We precomputed the optical flow using \cite{Liu2009Beyond} \footnote{\url{https://github.com/pathak22/pyflow}}.
%

\boldparagraph{SIP Subnet.}
Following \cite{Hui-et-al-CVPR-2018}, the SIP subnet is composed of 31 layers overall.
Two 3$\times$3 convolutional layers are first used to extract feature maps from the input -- in our case the output of the MVA subnet; followed by 4 distillation blocks, each consisting of an ``enhancement unit'' and a ``compression unit''.
A major difference to \cite{Hui-et-al-CVPR-2018} is that our SIP subnet operates directly in the target HR space. The bicubic upsampling is replaced by a skip connection, and instead of a final reconstruction block we use a 1$\times$1 convolution layer (see Figure~\ref{fig:network_architecture}).
Unsurprisingly, we found that upsampling feature maps at the end unavoidably introduces some blurring effect, whereas doing the enhancement of high-frequency content in the target domain is more effective.
We initialize all weights like \cite{he2015delving} and all biases to zero.

\subsection{Training Setup} \label{sec:training_setup}
%
Training was run on a GTX 1080 Ti GPU (12GB RAM).
The SIP subnet is first pre-trained on DIV2K data (batch size 64), then the complete MVA+SIP network is fine-tuned end-to-end on the 3D training set.
We use the Adam optimizer \cite{Diederik-et-al-ArXiv-2014} with learning rate $10^{-4}$.
Due to memory limitations, the batch size was set to 4 for the multi-view dataset.

\boldparagraph{Ground Truth Generation.}  An issue in texture
SR is the lack of ground truth, \ie, texture with a
higher resolution than that of the input images.
For each scene, we generated a pseudo-ground truth texture atlas by
running our MVA $L^1$ primal-dual model with an upsampling factor
$\times 2$ from all available views (\ie, using significantly more
viewpoints than for testing the complete network).
Since upsampling $\times 4$ in the same way leads to visually
imperfect results even with many views, we instead downsample the
input views to $\times 0.5$ their original size for experiments with
$\times 4$ super-resolution.
We note that the ground truth data may be biased towards the MVA subnet, since
it is generated with it, but for training and testing we use fewer and LR input views.

\begin{figure*}[t!]
  \centering
  \begin{scriptsize}
  \centering
  \newcommand{\sz}{1.25cm}
  \newcommand{\insz}{0.75cm}
  \newcommand{\rd}{5pt}
  \newcommand{\rulesep}{\unskip\ \vrule\ }
  \setlength{\tabcolsep}{2pt}
  \renewcommand{\arraystretch}{2}
  \vspace{-1em}
  \begin{tabular}{ccccccccccccccc}
  \multicolumn{7}{c}{\textbf{20-view model}}
  & &
  \multicolumn{7}{c}{\textbf{3-view model}} \\[-2mm]
  &
  \textit{Input atlas} & 
  \textit{MVA} & 
  \textit{MVA + SIP} & 
  \textit{GT} & 
  \textit{Res. MVA} &
  \textit{Res. SIP} &
  & 
  &
  \textit{Input atlas} & 
  \textit{MVA} & 
  \textit{MVA + SIP} & 
  \textit{GT} & 
  \textit{Res. MVA} &
  \textit{Res. SIP} \\[0pt]
  \rotatebox{90}{\hspace{5pt}\textit{Patch 1}} & 
  \includegraphics[height=\sz]{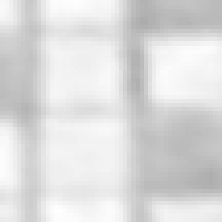} &
  \includegraphics[height=\sz]{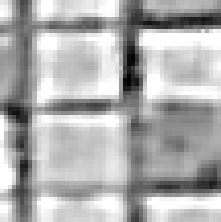} &
  \includegraphics[height=\sz]{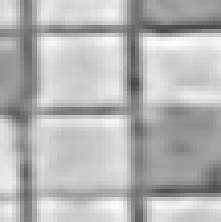} &
  \includegraphics[height=\sz]{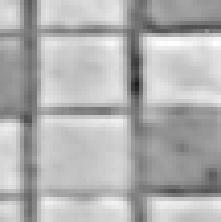} &
  \includegraphics[height=\sz]{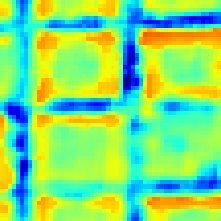} &
  \includegraphics[height=\sz]{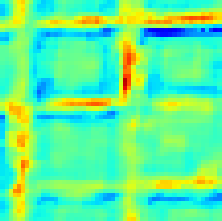}
    &
    & 
    \rotatebox{90}{\hspace{5pt}\textit{Patch 3}} &
  \includegraphics[height=\sz]{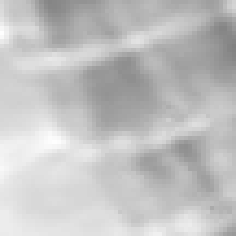} &
  \includegraphics[height=\sz]{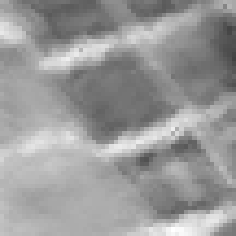} &
  \includegraphics[height=\sz]{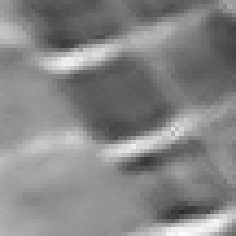} &
  \includegraphics[height=\sz]{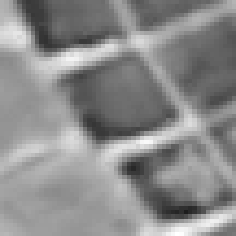} &
  \includegraphics[height=\sz]{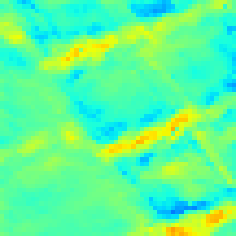} &
  \includegraphics[height=\sz]{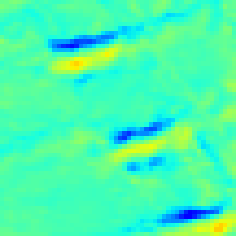}
  \\ 
  \rotatebox{90}{\hspace{5pt}\textit{Patch 2}} &      
  \includegraphics[height=\sz]{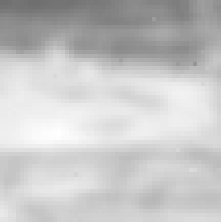} &
  \includegraphics[height=\sz]{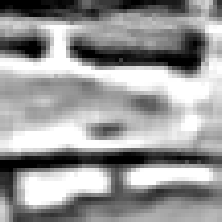} &
  \includegraphics[height=\sz]{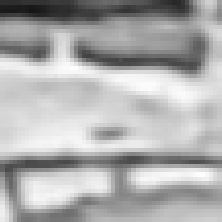} &
  \includegraphics[height=\sz]{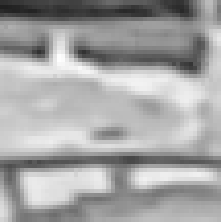} &
  \includegraphics[height=\sz]{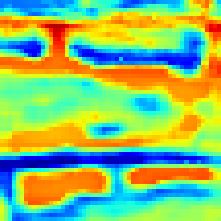} &
  \includegraphics[height=\sz]{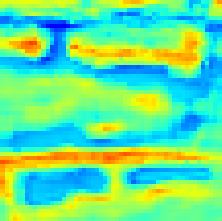}
  & 
  & 
  \rotatebox{90}{\hspace{5pt}\textit{Patch 4}} &
  \includegraphics[height=\sz]{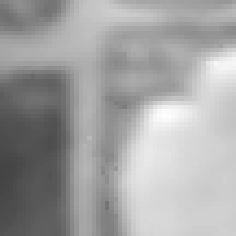} &
  \includegraphics[height=\sz]{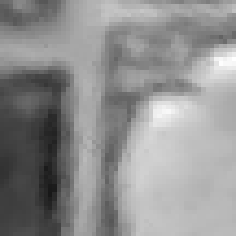} &
  \includegraphics[height=\sz]{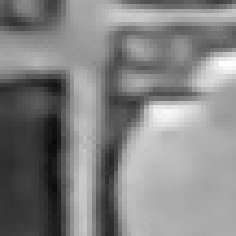} &
  \includegraphics[height=\sz]{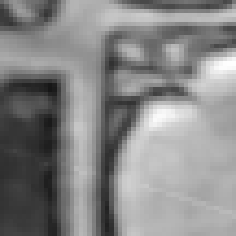} &
  \includegraphics[height=\sz]{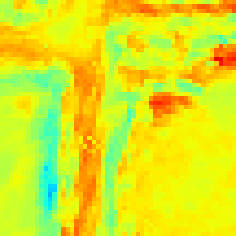} &
  \includegraphics[height=\sz]{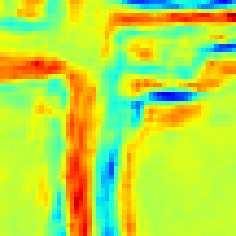} \\[-5pt]
  \end{tabular}
  \end{scriptsize}
  \caption{Texture patches at different steps of training
      (upsampling factor $\mathbf{\times 4}$). Our network takes as
    input a patch of the initial blurry atlas. The
    MVA subnet outputs a super-resolved patch, which is then
    further refined by the SIP subnet (MVA+SIP). Patches 1 and 2 show
    the 20-view case, patches 3 and 4 show the 3-view case.}
  \label{fig:training_patches}
\end{figure*}

\boldparagraph{Number of Input Views.}
At test time, the MVA subnet can process an arbitrary number of input views. For training, it is however convenient to have a fixed number. We trained two models, one with 20 and the other with 3 input views.
The purpose is to simulate two cases:
\begin{itemize}[topsep=1pt,itemsep=-1.5mm,leftmargin=*]
  \item 20 views emulate dedicated texture recording with healthy redundancy; thus favoring the MVA step.
  \item 3 views represent a more offhanded scenario with only few images; thus relying more on the SIP step.
\end{itemize} 
Figure~\ref{fig:training_patches} shows examples of training patches
for these two cases.  For the 20-view model the MVA subnet achieves
sharp super-resolution texture, but tends to exaggerate the intensity
contrast, the SIP subnet corrects the overshoot while preserving the
recovered details.
For the 3-view model, the MVA subnet can only partially recover the
fine structure, which is then further sharpened by the SIP subnet.
Overall, we find that the 3-view model generalizes better. This is
likely due to the fact that we had to generate our ground truth with a
HR, many-view version of MVA. As a consequence, a certain bias
towards MVA-type outputs is baked into the ground truth. Due to that
bias, MVA trained with many views gets ``too close'' to the imperfect
ground truth, which hampers the training of the SIP.  For the rest of
the paper, we run all tests with the model trained on 3 views, no
matter how many input views we feed it.

\boldparagraph{Runtime.}  Pre-training the SIP subnet on DIV2K takes
about 40 min per epoch (upsampling $\times 4$). Training time for the
full MVA+SIP takes 5 min per epoch for the 3-view model, respectively
20 min per epoch for the 20-view model.
At test time, super-resolving the largest texture atlas
(\emph{Fountain}) with upsampling factor $\times 4$ and 20 views takes
1 hour -- other scenes and settings are proportionally faster.
This is significantly faster than the most similar competing method
\cite{Tsiminaki-et-al-CVPR-2014}, which reports 30-60 minutes \emph{per
  iteration}.

\subsection{Results}
%
Our evaluation is divided into two parts. A comparison to
state-of-the-art methods at the upsampling factor $\times 2$ (the one
used in most previous work) and an ablation study at a more
interesting upsampling factor $\times 4$. 
%

\vspace{-0.4cm}
\subsubsection{Comparison with state-of-the-art}
\vspace{-0.1cm}
We compare our proposed end-to-end network with the state-of-the-art
multi-view texture SR techniques of
Tsiminaki~\etal\cite{Tsiminaki-et-al-CVPR-2014} and of
Goldl\"ucke~\etal \cite{Goldluecke-Cremers-DAGM-2009}.  We use the
same three objects \textit{Beethoven}, \textit{Bird} and
\textit{Bunny}, with identical views and 3D models.  The results for
competing methods are only available at upsampling factor $\times 2$,
thus we show results for that (rather moderate) super-resolution in
Figure~\ref{fig:sota_comparison}. We point out that, for reasons
unknown to us, the dataset made available to us differs from the
original one. Our version has 33, 20 and 36 views respectively for the
three test objects; whereas the literature suggests that the earlier
\cite{Tsiminaki-et-al-CVPR-2014} had access to up to 108 views.

Our approach delivers outputs with sharper details
than~\cite{Tsiminaki-et-al-CVPR-2014}, such as the letters and
eyebrows in \textit{Beethoven}, or the feathers of the \textit{Bird};
and without the over-sharpening artifacts
of~\cite{Goldluecke-Cremers-DAGM-2009}.
It also recovers the particularly challenging fur of the
\textit{Bunny}, although with little noticeable improvements over the
pure MVA method~\cite{Tsiminaki-et-al-CVPR-2014}.
We attribute this to the strong lighting differences between the input
views of the \emph{Bunny}.
As shown in Figure~\ref{fig:training_patches} the SIP subnet tries to
correct brightness deviations (cf.~Section~\ref{sec:training_setup}),
which may be problematic if the brightness varies across input
views.

Table~\ref{tab:sota_comparison} presents a quantitative comparison of
our method with two baselines (for an upsampling factor $\times 4$):
an re-implementation of \cite{Tsiminaki-et-al-CVPR-2014} with the more
robust $L^1$-dataterm, and the original version of the single-image
SR network \cite{Hui-et-al-CVPR-2018}.
It also shows respective performance of MVA, SIP and MVA+SIP network.
We report the SSIM, PSNR and SRE\footnote{\textit{Signal to
    Reconstruction Error}, measured as:
  $10\log_{10}\frac{\mu_{\mathbf{x}}^2}{\parallel \hat{\mathbf{x}} -
    \mathbf{x} \parallel^2 / n}$, with $\mu_{\mathbf{x}}$ the average
  of $\mathbf{x}$ and $n$ the number of pixels.}.  SSIM measures how
well the structures are recovered, while PSNR and SRE measure the
overall reconstruction error.  We can observe that in general our
network outputs better results than the two baselines. Looking at 
our MVA subnet alone, it achieves
high SSIM value, \ie, it recovers fine geometric structures. The
subsequent SIP mainly boosts the PSNR and SRE values, lifting them
significantly above prior art; corresponding to enhanced color and
contrast fidelity. While SIP on its own (without MVA) is not
competitive, showing its dependance on an already fairly high-quality
input. See also Section~\ref{discussion}.
%

\begin{table*}[ht]
\centering
 \begin{scriptsize}
  \setlength{\tabcolsep}{6.5pt}
  \begin{tabular}{l|ccc|ccc|ccc|ccc}
  \toprule
  & \multicolumn{3}{c|}{\textbf{Beethoven}} &  \multicolumn{3}{c|}{\textbf{Bird}}   & \multicolumn{3}{c|}{\textbf{Bunny}}   & \multicolumn{3}{c}{\textbf{Average}} \\
  &  \textit{SSIM}   &   \textit{PSNR}   &   \textit{SRE}    
  &  \textit{SSIM}   &   \textit{PSNR}   &   \textit{SRE}   
  &  \textit{SSIM}   &   \textit{PSNR}   &   \textit{SRE}   
  &  \textit{SSIM}   &   \textit{PSNR}   &   \textit{SRE}  \\
  \midrule 
  Tsiminaki \etal \cite{Tsiminaki-et-al-CVPR-2014} with $L^1$-dataterm      & 0.934	  & 40.488	 & 14.482   & 0.942	  & 41.379	  & 21.914  & 0.923  &	38.596   & 15.292  &  0.933	 & 40.155	& 17.229  \\
  \hdashline
  Hui \etal \cite{Hui-et-al-CVPR-2018} (pre-trained on \cite{Agustsson_2017_CVPR_Workshops})                       & 0.933	  & 39.505	 & 13.410   & 0.941	  & 41.999	  & 22.499  & 0.916	 & 37.888	 & 14.525  &  0.930	 & 39.797	& 16.811  \\
  \hline  
  MVA subnet                        & 0.931	  & 40.260	 & 14.253   & \textbf{0.944}	  & 41.901	  & 22.436  & 0.922	 & 38.573	 & 15.268  &  0.933	 & 40.245	& 17.319  \\
  \hdashline 
  SIP subnet (trained only on \cite{Agustsson_2017_CVPR_Workshops})  &  0.914  &	37.041	 &  2.009   &  0.921  & 38.738	  &  16.766  &  0.869 & 35.252	 &  7.006  &  0.901  & 37.010	&  8.594  \\
  \hdashline
  Ours (no pre-training, $\sigma_i$ fixed)                                 &  0.941  &	39.231	 &  13.224   &  0.940  &	39.146    &	19.680  &  0.929 & 38.198	 & 14.894  &  0.937	 & 38.858	& 15.933  \\
  Ours (SIP-Net pre-trained on 
  \cite{Agustsson_2017_CVPR_Workshops})   &  \textbf{0.948}  &	\textbf{43.309}	 & \textbf{17.304}   &  0.943  & \textbf{44.634}    & \textbf{25.171} &  \textbf{0.932} &  \textbf{39.690}   & \textbf{16.386}  &  \textbf{0.941}	 & \textbf{42.544}   & \textbf{19.620}  \\
  \bottomrule
  \end{tabular}
\end{scriptsize}
  \caption{Quantitative comparison of different texture
    super-resolution techniques (upscaling factor $\times4$, same
    initial texture atlas, all computed on the Y-channel images). 
    The top two rows are baselines : 
    our reimplementation of \cite{Tsiminaki-et-al-CVPR-2014} with
    primal-dual optimization scheme and $L^1$-dataterm (first row),
    single-image super-resolution network \cite{Hui-et-al-CVPR-2018} 
    (second row). We evaluate the individual components of our proposed
    approach: MVA subnet trained on our data, which is very similar to
    \cite{Tsiminaki-et-al-CVPR-2014}, but estimates the local blur
    from the data (third row); SIP subnet alone trained only on DIV2K (fourth row);
    and our complete network MVA+SIP (last two rows).
   }
  \label{tab:sota_comparison}
\end{table*}

\begin{figure*}[t!]
  \centering
  \setlength{\tabcolsep}{0.5mm}
  \begin{scriptsize}
  \centering
  \newcommand{\sz}{2.05cm}
  \newcommand{\insz}{1.5cm}
  \newcommand{\rd}{1pt}
  \begin{tabular}{cccccccccc}
  \textit{3D Model} & 
  \textit{Input} & 
  \cite{Goldluecke-Cremers-DAGM-2009} & 
  \cite{Tsiminaki-et-al-CVPR-2014} &
  \textit{Ours} 
  & &
  \textit{Input} & 
  \cite{Goldluecke-Cremers-DAGM-2009} & 
  \cite{Tsiminaki-et-al-CVPR-2014} &
  \textit{Ours} 
\\[2pt]
  \includegraphics[height=\sz]{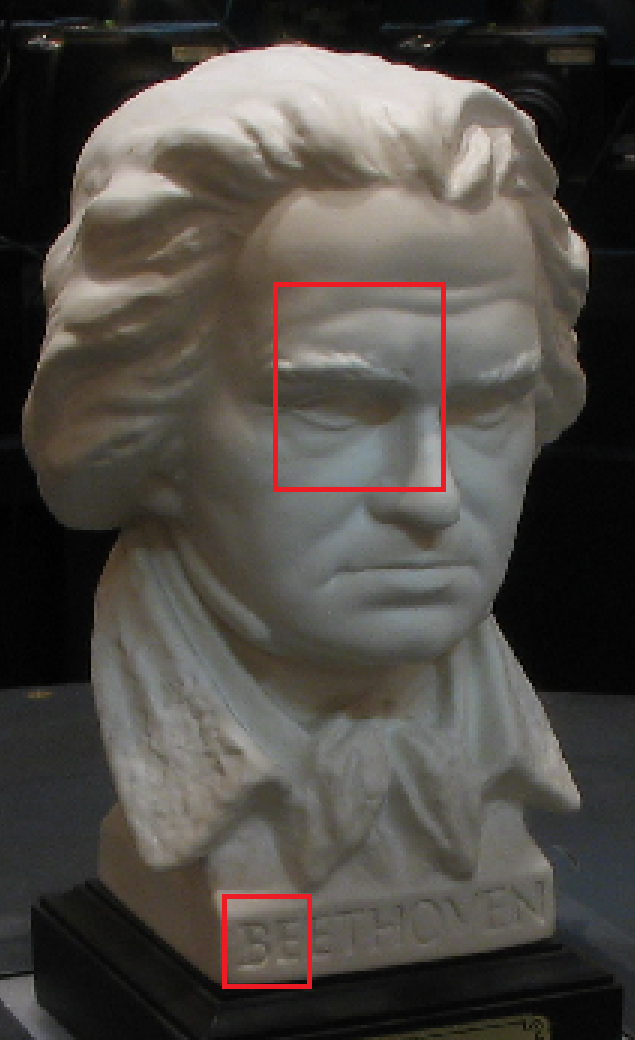} &
  \includegraphics[height=\sz]{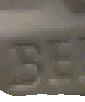} &
  \includegraphics[height=\sz]{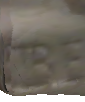} &
  \includegraphics[height=\sz]{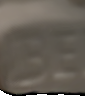} &
  \includegraphics[height=\sz]{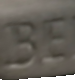} & &
  \vspace{\rd}
  \includegraphics[height=\sz]{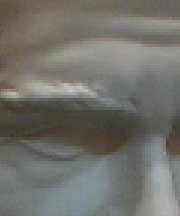} &
  \includegraphics[height=\sz]{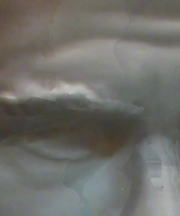} &
  \includegraphics[height=\sz]{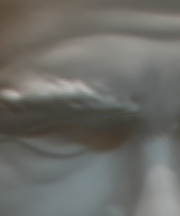} &
  \includegraphics[height=\sz]{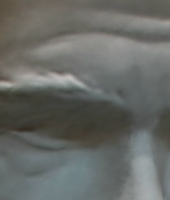} 
  \\
  \includegraphics[height=\sz]{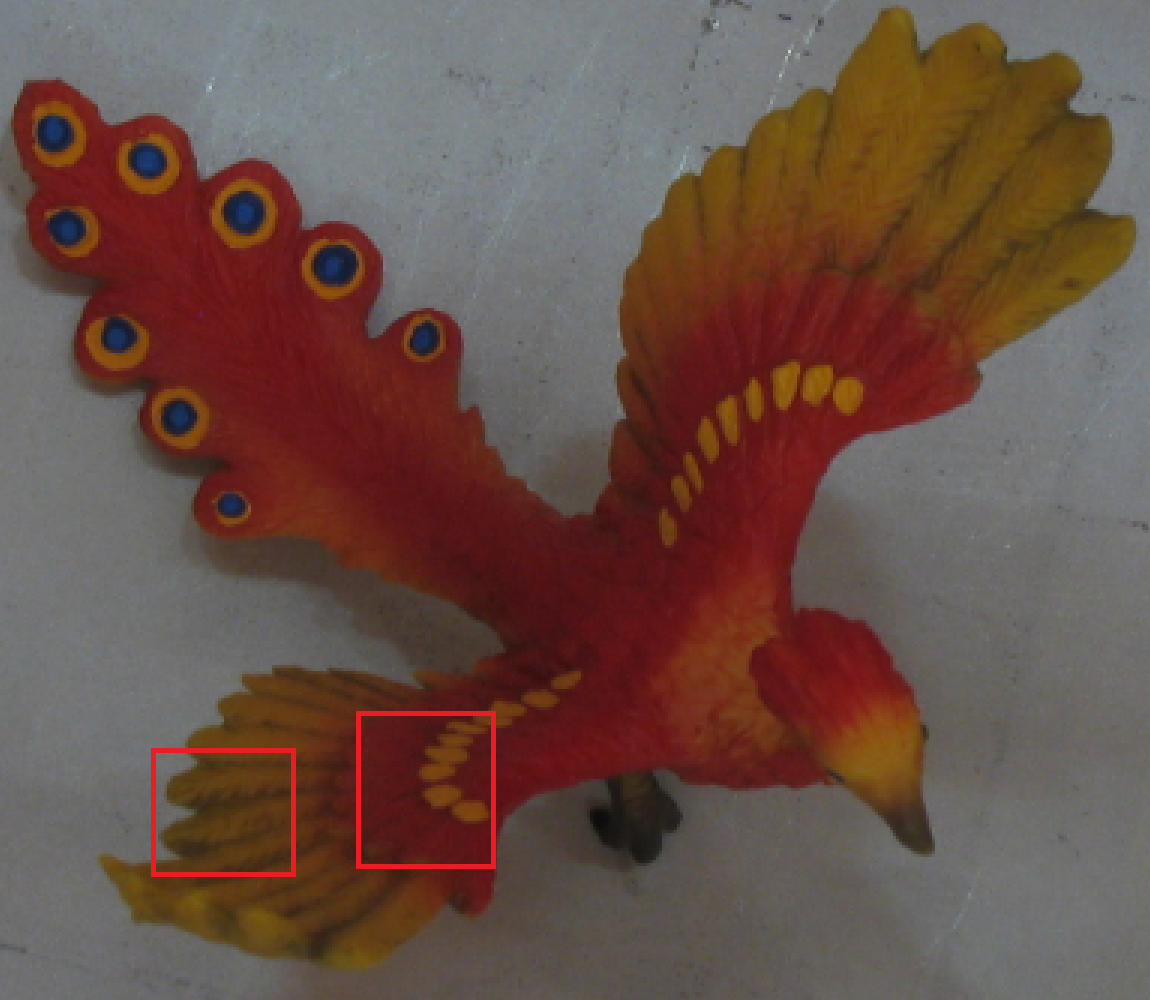}&
  \includegraphics[height=\sz]{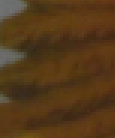} &
  \includegraphics[height=\sz]{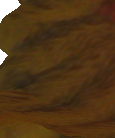} &
  \includegraphics[height=\sz]{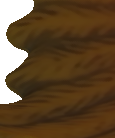} &
  \includegraphics[height=\sz]{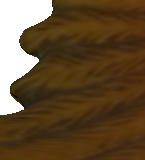} & &
  \vspace{\rd}
  \includegraphics[height=\sz]{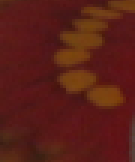} &
  \includegraphics[height=\sz]{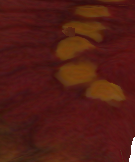} &
  \includegraphics[height=\sz]{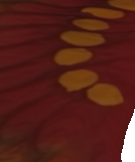} &
  \includegraphics[height=\sz]{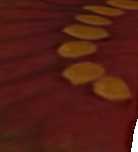} 
\\
  \includegraphics[height=\sz]{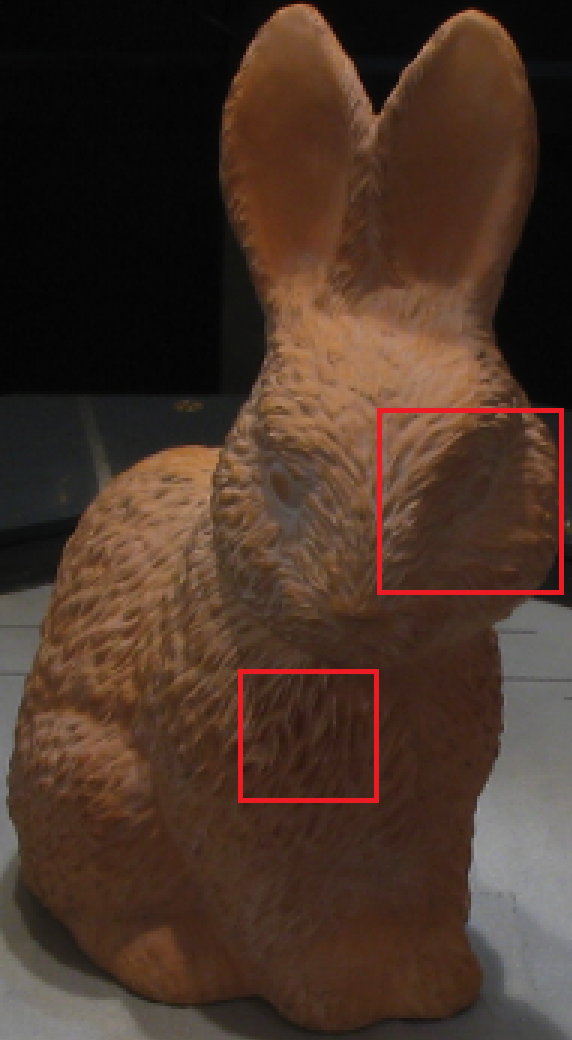} &
  \includegraphics[height=\sz]{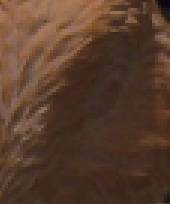} &
  \includegraphics[height=\sz]{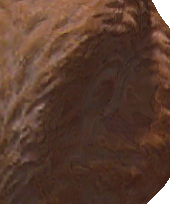} &
  \includegraphics[height=\sz]{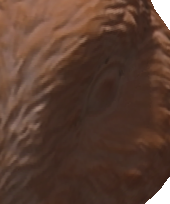} &
  \includegraphics[height=\sz]{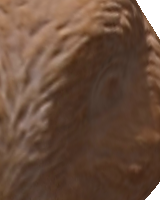} & &
  \vspace{\rd}
  \includegraphics[height=\sz]{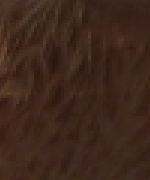} &
  \includegraphics[height=\sz]{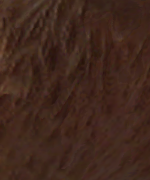} &
  \includegraphics[height=\sz]{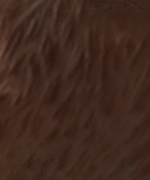} &
  \includegraphics[height=\sz]{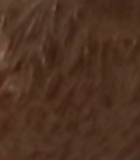} \\[-1mm]

  \end{tabular}
  \end{scriptsize}
  \caption{Qualitative comparison to state-of-the-art
      multi-view SR methods (upscaling factor $\times 2$). We
    compare to~\cite{Goldluecke-Cremers-DAGM-2009}
    and~\cite{Tsiminaki-et-al-CVPR-2014} for which the authors
    provided results for \textit{Beethoven}, \textit{Bird} and
    \textit{Bunny} datasets (computed from 108, 52 and 52 input
    images, respectively). Our results were obtained with the
    available LR views, respectively 33, 20 and 36.}
  \label{fig:sota_comparison}
  \vspace{-0.15cm}
\end{figure*}

\vspace{-0.15cm}
\subsubsection{Ablation study}
\begin{figure*}[t!]
  \vspace{-0.2cm}
  \centering
  \begin{scriptsize}
  \centering
  \newcommand{\sz}{2.1cm} 
  \newcommand{\insz}{1.75cm}
  \newcommand{\beethovenheight}{1.4cm} 
  \newcommand{\birdheight}{1.6cm} 
  \newcommand{\bunnyheight}{1.6cm} 
  \newcommand{\rd}{5pt}
  \setlength{\tabcolsep}{1pt}
   
  \begin{tabular}{ccccccccc}
  & \textit{3D Model} & \textit{Input View} & \textit{GT} & \textit{Ours} &
    \textit{MVA} & \textit{SIP} & \textit{SR + \cite{Waechter-et-al-ECCV-14}} &
    \textit{\cite{Waechter-et-al-ECCV-14} from LR} \\
  \multirow{ 2}{*}{\rotatebox{90}{\hspace{12pt}\textit{Beethoven}}} & 
  \includegraphics[height=\beethovenheight, width=\insz]{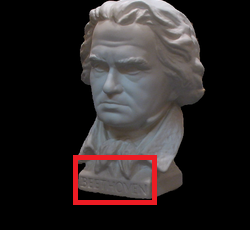} &
  \includegraphics[height=\beethovenheight, width=\sz]{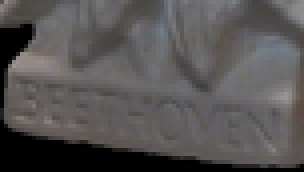} &
  \includegraphics[height=\beethovenheight, width=\sz]{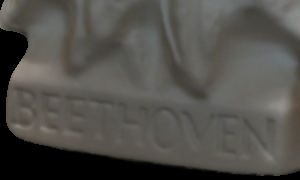} &
  \includegraphics[height=\beethovenheight, width=\sz]{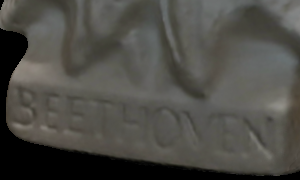} &
  \includegraphics[height=\beethovenheight, width=\sz]{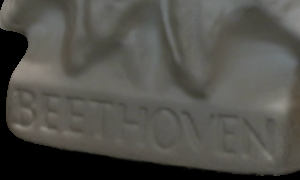} &
  \includegraphics[height=\beethovenheight, width=\sz]{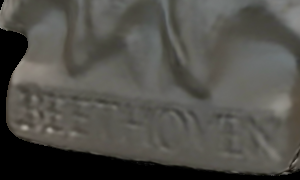} &
  \includegraphics[height=\beethovenheight, width=\sz]{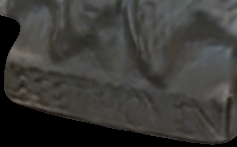} &
  \includegraphics[height=\beethovenheight, width=\sz]{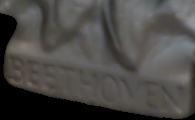} \\
  &
  \includegraphics[height=\beethovenheight, width=\insz]{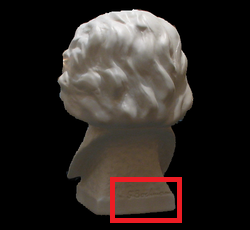} 
  &  
  \includegraphics[height=\beethovenheight, width=\sz]{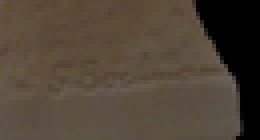} &
  \includegraphics[height=\beethovenheight, width=\sz]{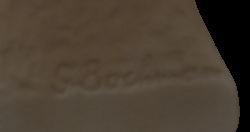} &
  \includegraphics[height=\beethovenheight, width=\sz]{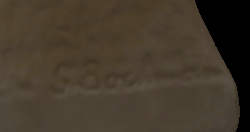} &
  \includegraphics[height=\beethovenheight, width=\sz]{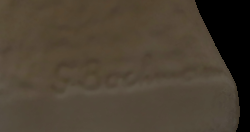} &
  \includegraphics[height=\beethovenheight, width=\sz]{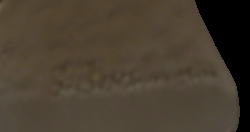} &
  \includegraphics[height=\beethovenheight, width=\sz]{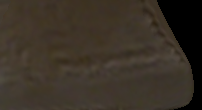} &
  \includegraphics[height=\beethovenheight, width=\sz]{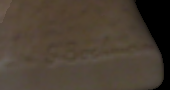} \\
  \multirow{ 2}{*}{\rotatebox{90}{\hspace{12pt}\textit{Bird}}} &
  \includegraphics[height=\birdheight, width=\insz]{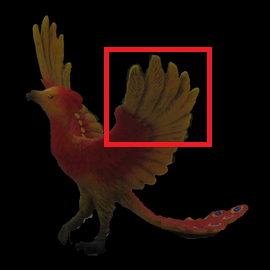} &  
  \includegraphics[height=\birdheight, width=\sz]{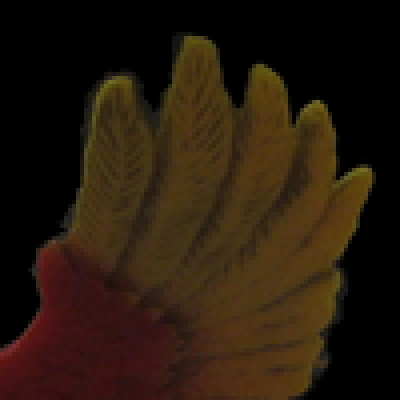} &
  \includegraphics[height=\birdheight, width=\sz]{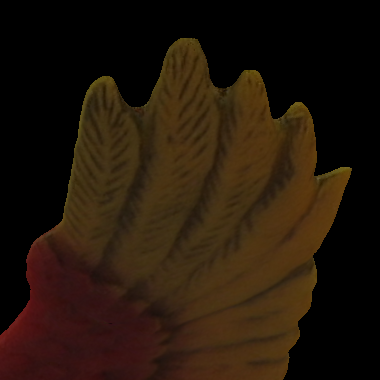} & 
  \includegraphics[height=\birdheight, width=\sz]{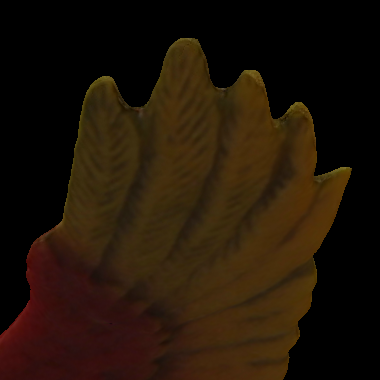} &
  \includegraphics[height=\birdheight, width=\sz]{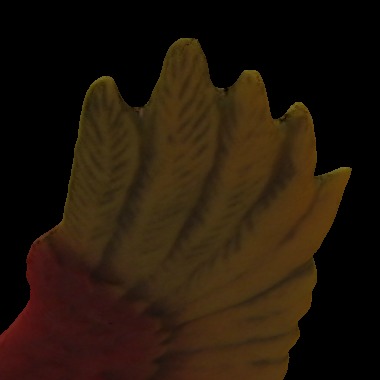} &
  \includegraphics[height=\birdheight, width=\sz]{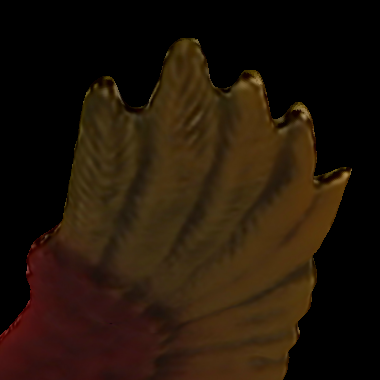} &
  \includegraphics[height=\birdheight, width=\sz]{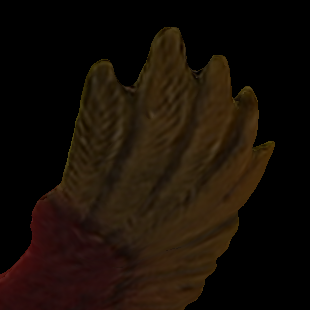} &
  \includegraphics[height=\birdheight, width=\sz]{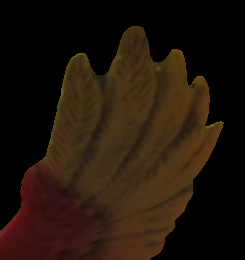} \\
  &
  \includegraphics[height=\birdheight, width=\insz]{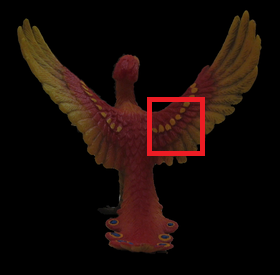} &  
  \includegraphics[height=\birdheight, width=\sz]{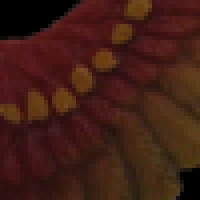} &
  \includegraphics[height=\birdheight, width=\sz]{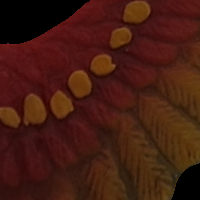} &
  \includegraphics[height=\birdheight, width=\sz]{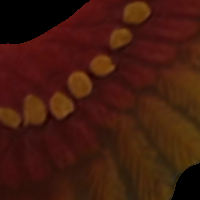} &
  \includegraphics[height=\birdheight, width=\sz]{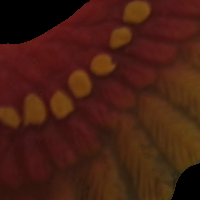} &
  \includegraphics[height=\birdheight, width=\sz]{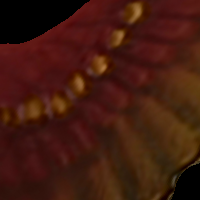} &
  \includegraphics[height=\birdheight, width=\sz]{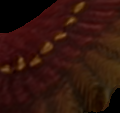} &
  \includegraphics[height=\birdheight, width=\sz]{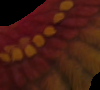} \\
  \multirow{ 2}{*}{\rotatebox{90}{\hspace{12pt}\textit{Bunny}}} &
  \includegraphics[height=\bunnyheight, width=\insz]{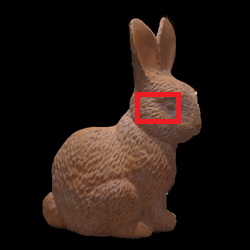} &  
  \includegraphics[height=\bunnyheight, width=\sz]{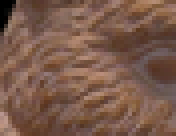} &
  \includegraphics[height=\bunnyheight, width=\sz]{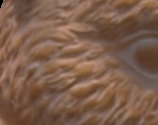} &
  \includegraphics[height=\bunnyheight, width=\sz]{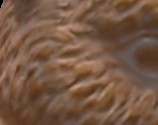} &
  \includegraphics[height=\bunnyheight, width=\sz]{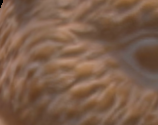} &
  \includegraphics[height=\bunnyheight, width=\sz]{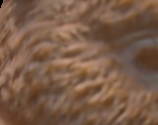} &
  \includegraphics[height=\bunnyheight, width=\sz]{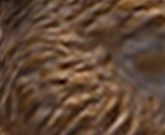} &
  \includegraphics[height=\bunnyheight, width=\sz]{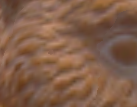} \\
  &
  \includegraphics[height=\bunnyheight,width=\insz]{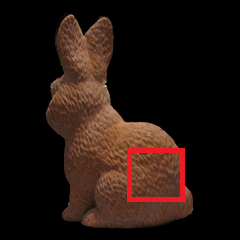} &  
  \includegraphics[height=\bunnyheight, width=\sz]{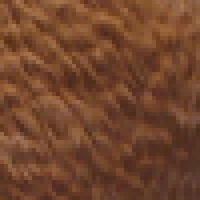} &
  \includegraphics[height=\bunnyheight, width=\sz]{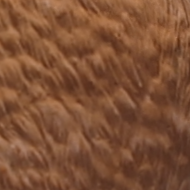} &
  \includegraphics[height=\bunnyheight, width=\sz]{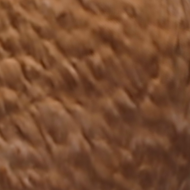} &
  \includegraphics[height=\bunnyheight, width=\sz]{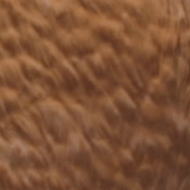} &
  \includegraphics[height=\bunnyheight, width=\sz]{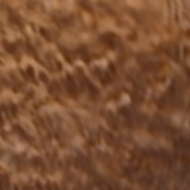} &
  \includegraphics[height=\bunnyheight, width=\sz]{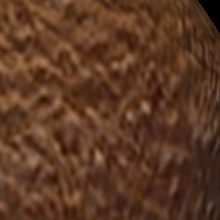} &
  \includegraphics[height=\bunnyheight, width=\sz]{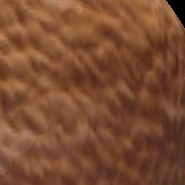} \\[1mm]  
  \end{tabular}
  \end{scriptsize}
  \caption{Ablation study and comparison (upsampling factor $\times 4$). We compare our result (fourth column) to: LR input view                	(second column), ground truth obtained from \cite{Tsiminaki-et-al-CVPR-2014} with upsampling factor $\times 2$ from the original images
(third column); output of the MVA subnet (fifth column); output of the SIP subnet without MVA (sixth column); output texture of 
\cite{Waechter-et-al-ECCV-14} from input views super-resolved with \cite{Hui-et-al-CVPR-2018} (seventh column); and output texture directly from LR input views with \cite{Waechter-et-al-ECCV-14} (eighth column).}
  \label{fig:ablation}
  \vspace{-0.3cm}
\end{figure*}

We also conducted an ablation study to assess the contribution of each
subnet, see Figure~\ref{fig:ablation}. For this experiment the textures
were super-resolved with a bigger upsampling factor of $\times 4$
to challenge our method.

As expected, generating a texture atlas from LR images with a state-of-the-art
atlas generator such as \cite{Waechter-et-al-ECCV-14} leads to LR results. 
Moreover, independently super-resolving the input views and
using them as input to the same generator \cite{Waechter-et-al-ECCV-14} may lead to
misalignments. 
The single-image SR method may hallucinate different high-frequency details in different views which are then hard to align.
This is particularly visible on the close-ups from the \emph{Beethoven}
dataset, where the letters in \textit{''BEETHOVEN''} and
\textit{''S. BOCHMANN''} are not correctly aligned.  
Leveraging the redundant information from multiple input views, the
MVA subnet can recover sharp details, such as the fur of the
\emph{Bunny}, the feathers of the \emph{Bird}, or the letters on
\emph{Beethoven}.  The SIP subnet on the other hand excels at
enhancing existing details, cf.~Table~\ref{tab:sota_comparison}.
Comparing the output of the MVA subnet with the full approach, we can
see that the letters and stone texture in \emph{Beethoven} become
sharper, as does the fur around the eye of the \emph{Bunny}.
Moreover, the output of SIP without preceding MVA confirms the
mentioned sensitivity of SIP to its initialization.

\vspace{-0.4cm}
\subsubsection{Discussion}
\label{discussion}
\vspace{-0.1cm}
\boldparagraph{Training Data.}
It was a challenge to find sufficient training data for our method. In
particular, to train the MVA subnet one needs not only a textured 3D
model, but also the projection operators $P_i$ of the $N$ original views used
to create the texture.
Due to the scarcity of such datasets, we could not afford to set aside
more than three scenes (\emph{Beethoven}, \emph{Bird} and
\emph{Bunny}), in order to have sufficient training data.
As further visual examples on more varied scenes we also show results
obtained on some of the training scenes (\emph{Fountain},
\emph{Buddha} and \emph{Relief}) in Figure~\ref{fig:teaser} and in
\textit{Supplementary Material}. According to good practice in machine
learning, these results were not used for quantitative evaluation.

\boldparagraph{Texture Initialization.}
In our experiments we found that the prior-based SIP subnet is 
sensitive to the quality of the LR input texture.
This can be observed in Figure~\ref{fig:ablation} and Table~\ref{tab:sota_comparison}.
If the input atlas is too blurry, e.g., due
to averaging of imperfectly aligned images, then the redundancy-based
MVA subnet is still able to super-resolve the patch. Whereas the SIP
subnet has trouble to add further high-frequency details.
On the contrary, if the input atlas already contains fine details, the MVA might not improve much, 
but the SIP subnet manages to further enhance them.

The MVA subnet approximates a variational
optimization of a convex energy and is therefore
independent of the initialization; whereas the SIP subnet performs a
small, and at most locally optimal, residual correction.
Overall, the results confirm our intuition that the MVA and SIP are to
some degree complementary and that their combination can achieve
superior SR.

\section{Conclusion}
We presented a novel multi-view texture super-resolution network
that unifies and exploits two fundamental approaches to
super-resolution, namely physically motivated multi-view SR and
prior-based single-view SR.  Our end-to-end neural network design
combines modern deep learning techniques with classical energy
minimization methods via optimization unrolling.  This leads to a
problem-specific network architecture which avoids the learning of the
perspective projection operator.  Further, we can handle a varying
number of input images and are invariant to their ordering.  Several
experiments demonstrate that the approach outperforms prior art in
texture SR.

{\scriptsize
\vspace{1em}
\boldparagraph{Acknowledgements.}
This research was partially supported by the Intelligence Advanced Research Projects Activity (IARPA) via Department of Interior/ Interior Business Center (DOI/IBC) contract number D17PC00280. The U.S. Government is authorized to reproduce and distribute reprints for Governmental purposes notwithstanding any copyright annotation thereon. Disclaimer: The views and conclusions contained herein are those of the authors and should not be interpreted as necessarily representing the official policies or endorsements, either expressed or implied, of IARPA, DOI/IBC, or the U.S. Government. 
\par
}

\clearpage

{\small
\bibliographystyle{ieee}
\bibliography{bibliography}
}

\end{document}